\documentclass[10pt,twocolumn,letterpaper]{article}

\usepackage{cvpr}              

\usepackage{graphicx}
\usepackage{amsmath}
\usepackage{amssymb}
\usepackage{booktabs}

\usepackage{epsfig}
\usepackage{graphics}
\usepackage{subfloat}
\usepackage{color}
\usepackage{mathtools}
\usepackage{multirow}
\usepackage{tabularx}
\usepackage{amsfonts}
\usepackage{wrapfig,lipsum,booktabs}
\usepackage{xcolor}
\usepackage{floatflt}
\usepackage[cal=cm]{mathalfa} 
\usepackage[pagebackref=true,breaklinks=true,letterpaper=true,colorlinks,bookmarks=false]{hyperref}
\usepackage[ruled, lined, linesnumbered, commentsnumbered, longend]{algorithm2e}

\usepackage{caption}
\usepackage{enumitem}
\setitemize{noitemsep,topsep=0pt,parsep=0pt,partopsep=0pt}
\usepackage{pifont}
\newcommand{\cmark}{\ding{51}}%
\newcommand{\xmark}{\ding{55}}%

\def\eg{\textit{e.g.,}}
\def\ie{\textit{i.e.,}}

\def\etal{\textit{et al.}}

\def\cvprPaperID{863} 
\def\confName{CVPR}
\def\confYear{2022}

\begin{document}

\title{Global and Local Alignment Networks for Unpaired Image-to-Image Translation}

\author{Guanglei Yang$^{1,2}$  \quad  Hao Tang$^3$ \quad Humphrey Shi$^{4,5,6}$  \quad Mingli Ding$^1$  \\
 Nicu Sebe$^2$ \quad Radu Timofte$^3$ \quad Luc Van Gool$^3$  \quad  Elisa Ricci$^{2,7}$\\
	$^1$Harbin Institute of Technology  \quad $^2$DISI, University of Trento \quad $^3$Computer Vision Lab, ETH Zurich \\ $^4$University of Oregon \quad $^5$UIUC  \quad $^6$Picsart AI Research (PAIR) \quad $^7$Fondazione Bruno Kessler
}

\maketitle

\begin{abstract}

The goal of unpaired image-to-image translation is to produce an output image reflecting the target domain's style while keeping unrelated contents of the input source image unchanged.
However, due to the lack of attention to the content change in existing methods, the semantic information from source images suffers from degradation during translation. 
In the paper, to address this issue, we introduce a novel approach, Global and Local Alignment Networks (GLA-Net).
The global alignment network aims to transfer the input image from the source domain to the target domain. To effectively do so, we learn the parameters (mean and standard deviation) of multivariate Gaussian distributions as style features by using an MLP-Mixer based style encoder.
To transfer the style more accurately, we employ an adaptive instance normalization layer in the encoder, with the parameters of the target multivariate Gaussian distribution as input. We also adopt regularization and likelihood losses to further reduce the domain gap and produce high-quality outputs. Additionally, we introduce a local alignment network, which employs a pretrained self-supervised model to produce an attention map via a novel local alignment loss, ensuring that the translation network focuses on relevant pixels.
Extensive experiments conducted on 
five public datasets demonstrate that our method effectively generates sharper and more realistic images than existing approaches.
Our code is available at \url{https://github.com/ygjwd12345/GLANet}.

\end{abstract}
\section{Introduction}
Recently, Generative Adversarial Networks (GANs)~\cite{goodfellow2014generative} have enabled the construction of powerful deep networks
for domain adaptation~\cite{dai2018dark,sakaridis2019guided} and image-to-image translation~\cite{huang2017arbitrary,kim2016accurate,gatys2016image,dong2015image}. 
Focusing on image-to-image translation, most methods assume paired images among domains. However, for many tasks, pairwise information is hard to get and often not available.
To tackle this problem, several works (\ie~CycleGAN~\cite{zhu2017unpaired}, DiscoGAN~\cite{kim2017learning}, DualGAN~\cite{yi2017dualgan}, and StarGAN~\cite{choi2020stargan}) proposed to employ a cycle consistency loss in order to transfer images from the source to the target domain without pairwise supervision.
While effective, these approaches suffer from performance degradation when significant changes must be operated to transfer an image to the other domain. Indeed,  most existing methods can successfully transfer low-level information, such as color or texture, but struggle to control information changes at the semantic level.

\begin{figure}[!t]
\subfloat[Input Image]{
    \begin{minipage}{0.4\linewidth}
        \centering
        \includegraphics[width=0.993\textwidth,height=1in]{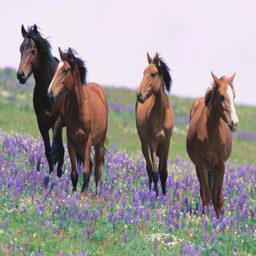}\\
    \end{minipage}%
}
\subfloat[U-GAT-IT~\cite{kim2020u}]{
    \begin{minipage}{0.4\linewidth}
        \centering
        \includegraphics[width=0.993\textwidth,height=1in]{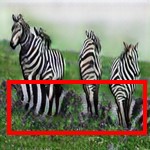}\\
    \end{minipage}%
}\hfil
\subfloat[AttentionGAN~\cite{tang2021attentiongan}]{
    \begin{minipage}{0.4\linewidth}
        \centering
        \includegraphics[width=0.993\textwidth,height=1in]{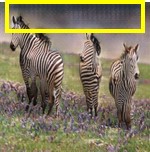}\\
    \end{minipage}%
}
\subfloat[GLA-Net (Ours)]{
    \begin{minipage}{0.4\linewidth}
        \centering
        \includegraphics[width=0.993\textwidth,height=1in]{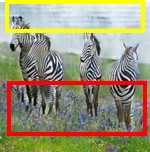}\\
    \end{minipage}%
}%
\centering
\caption{Comparison with state-of-the-art unsupervised image translation methods (\ie~AttentionGAN~\cite{tang2021attentiongan} and U-GAT-IT~\cite{kim2020u}) on an example of Horse$\to$Zebra task. The difference between the proposed GLA-Net and AttentionGAN~\cite{tang2021attentiongan} is marked with a yellow box while the difference between GLA-Net and U-GAT-IT~\cite{kim2020u} is marked with a red box. Better seen with magnification.}
\label{fig:teaser}
\vspace{-0.4cm}
\end{figure}

To solve this problem, ContrastGAN~\cite{liang2018generative} proposed a novel mask-conditional adversarial contrasting loss to enforce the produced output images to be semantically closer to the real data within the target category. However, this method requires object mask annotations in the training phase, thus implying a time-consuming labeling process. 
Similarly, Attention-GAN~\cite{chen2018attention} introduced an extra attention network to produce a spatial attention map to distinguish among regions of interest and background and facilitate image generation. Following~\cite{chen2018attention}, Tang~\etal~\cite{tang2021attentiongan} proposed to integrate the attention network into the generator to identify foreground objects and only modify them while keeping the background unchanged.  U-GAT-IT~\cite{kim2020u} introduced adaptive layer-instance normalization to guide the attention-based models. 
However, the main problem with all these methods is that style and content change are not considered simultaneously. In other words, while these approaches either apply attention maps or instance normalization to control the transfer of semantics and style, 
{they often fail to take care of both, the main reason being that style and content cannot be regarded as independent factors.}
An example of this phenomenon is illustrated in Figure~\ref{fig:teaser}. While U-GAT-IT~\cite{kim2020u} successfully manages to replace the horse style with the zebra style, there is still content degradation in the area of the image near the horse (red box in Figure~\ref{fig:teaser} (b)). Differently, in AttentionGAN~\cite{tang2021attentiongan}, while the attention map correctly forces the network to focus on the important areas of the image, some artifacts are visible in the style transition part (yellow box in Figure~\ref{fig:teaser} (c)).

\begin{figure*}[t] \small
\centering
\includegraphics[width=0.91\textwidth]{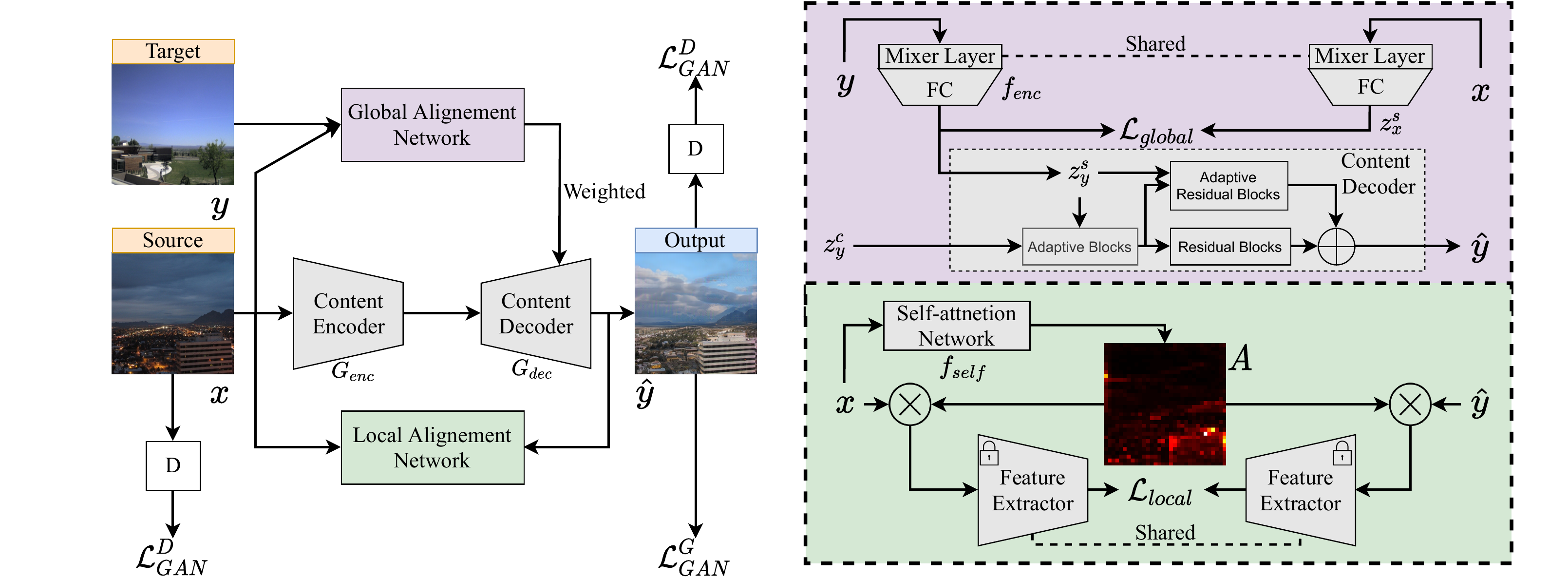}
\caption{Overview of GLA-Net. The whole architecture is composed by a global alignment network, a local alignment network, a content encoder $G_{enc}$, and a content decoder $G_{dec}$. In the global alignment network, we employ a style encoder $f_{enc}$  with different fully-connected layers but shared-weight MLP-Mixer layers to extract style features, $z_x^s$ and $z_y^s$, from source image $x$ and target image $y$. The global alignment loss $\mathcal{L}_{global}$ aims to make style feature between the two domain as close as possible.
The style features of the target image $z_y^s$, including $\mu_y,\sigma_y$, are also used to assign weights and biases to the adaptive blocks of the content decoder. As for the local alignment network, we adopt a self-attention network $f_{self}$  to get an attention map $A$ which weights source image and output image, and then is fed to the local alignment loss $\mathcal{L}_{local}$.
{In this way, it can help the network to avoid content destruction in unrelevant part.} The symbol \textcircled{x} and \textcircled{+} denotes element-wise multiplication and element-wise addition, respectively.}
\label{fig:overview}
\vspace{-0.4cm}
\end{figure*}

In this paper, we advance the state-of-the-art in unpaired image-to-image translation by proposing a novel framework, Global and Local Alignment Networks (\ie~GLA-Net), which seamlessly addresses the subtasks of style transfer and semantic change (see Figure \ref{fig:overview}).
Specifically, in the global alignment network, we use the style features predicted by our network to replace the learnable scale and shift parameter in instance normalization, thus implementing a novel and more effective style transform. Moreover, we introduce a global alignment loss to enforce similarities among the two multivariate Gaussian distributions obtained from the source and target domains.
Since the global alignment network is designed to account for style transfer, the semantic information of unrelated pixels or backgrounds suffers from degradation during image translation in the case of significant changes. To support the global alignment network,
we further propose a novel local alignment network. It helps the generator to focus on essential pixels during image translation and reduce the content degradation in irrelevant image parts as much as possible.
Thanks to the combination between the global and the local alignment network, the generator is able to successfully operate a global style transfer and produce consistent content changes.

\noindent The contributions of this paper are summarized as follows:
\begin{itemize}[leftmargin=*]
    \item We propose a novel framework for unpaired image-to-image translation which simultaneously addresses style transfer and semantic content change tasks within the same deep model, thus enabling higher accuracy and flexibility in style and content modification. 
    \item  We introduce GLA-Net, a novel deep architecture with two main components: a global and a local alignment network. The global alignment network enables accurate style transfer, thanks to an MLP-Mixer style encoder and a feature alignment strategy.
    The local alignment network integrates a self-supervised attention map to mitigate the content degradation problem of existing methods.
    \item Extensive experiments conducted on five publicly available datasets demonstrate that GLA-Net can generate photo-realistic images with higher quality and sharper details than existing methods.
\end{itemize}
\section{Related Work}

Image-to-image translation methods~\cite{lee2018diverse,liu2019few,tang2019attention,wang2018high,zhu2017unpaired} operate by transferring the style of images from a source to a target domain while preserving the content information.  
Earlier methods~\cite{chen2017photographic,isola2017image,park2019semantic,wang2018high} mainly focused on paired image-to-image translation, thus heavily relying on a reconstruction loss computed on paired images among domains.
However, in many real-world applications, pairwise information is not available.   
To address this issue, several works considered a cycle consistency loss in order to preserve the consistency at image level \cite{zhu2017unpaired,kim2017learning,yi2017dualgan} or at feature level \cite{zhu2017multimodal,huang2018multimodal}.
However, as the cycle consistency loss treats all pixels equally, these models struggle to 
focus on the essential parts of the images. 
To solve this problem, attention maps can be integrated to guide the translation model similar to other computer vision tasks such as depth estimation ~\cite{xu2018structured,yang2021transformer} and semantic segmentation~\cite{fu2019dual,hu2018squeeze}. 
Two strategies are possible to create attention modules that compute the region of interest for the image translation task. The first strategy consists in using extra data to provide attention. For instance, ContrastGAN~\cite{liang2018generative} used object mask annotations as extra input data.  InstaGAN~\cite{mo2019instagan} also proposed to incorporate the object segmentation mask to improve multi-instance transfiguration. The second strategy is to train another segmentation or attention model to generate attention maps and insert them into the system. This strategy was considered in~\cite{yang2019show}, SPADE~\cite{park2019semantic}, SEAN~\cite{zhu2020sean} and AttentionGAN~\cite{tang2021attentiongan}. 

Another problem with earlier approaches for unsupervised image-to-image translation~\cite{zhu2017unpaired,kim2017learning} is that, only with cycle consistency loss, the local semantic information tends to be destroyed in image-to-image translation, thus affecting the overall quality of the generated image. To address this issue, Park \etal~\cite{park2020contrastive} proposed CUT and applied contrastive learning to learn the correspondence between associated patches.
Similarly, FSeSim~\cite{zheng2021spatially} introduced a spatially-correlative loss to capture
patch-wise spatial relationships within an image.
In this paper, we introduce a different strategy to address this problem and propose a novel local alignment network that integrates a self-attention mechanism. 

Ensuring effective style transfer is a fundamental problem in image-to-image translation. Style transfer models typically operate by normalizing feature tensors with instance-specific mean and standard deviation, \emph{i.e.}, adopt Instance Normalization (IN)\cite{ulyanov2016instance,dumoulin2017learned}. %
Adaptive Instance Normalization (AdaIN)~\cite{huang2017arbitrary} represent an improvement over traditional IN. 
In image-to-image translation, to ensure better style transfer among domains, U-GAT-IT~\cite{kim2020u} proposes an adaptive layer-instance normalization, whose parameters are learned from datasets during the training stage. 
Different from U-GAT-IT~\cite{kim2020u}, in this paper, we propose a distribution alignment approach to extend AdaIN and integrate it into our global alignment network.
Furthermore, while all the above methods focus on specific aspects related either to style transfer or semantic content modification, in this paper, we propose for the first time an architecture that simultaneously handles the style and content changes in a holistic manner.

\section{Global and Local Alignment Networks}

In this paper, we present a novel framework, \ie~Global and Local Alignment Networks (GLA-Net), to simultaneously realize style transfer and semantic content modification for unpaired image-to-image translation. 
In the following, we first introduce the whole framework architecture in Section~\ref{sec:framework}. Then, in Section~\ref{sec:global} and Section~\ref{sec:local}, we introduce the details of our GLA-Net. Finally, we discuss the loss functions used in our framework in Section~\ref{sec:opt}.

\subsection{Framework Overview}
\label{sec:framework}

Given a collection of images $\mathcal{X}{\subset} \mathbb{R}^{H\times W\times C}$ from a source domain, our main goal is to learn a model $G$ that receives the image $x$ ${\in} \mathcal{X}$ as input and transfers it into the target domain $\mathcal{Y}{\subset} \mathbb{R}^{H\times W\times C}$, in a way to jointly convert the style and semantic content together. 

Figure~\ref{fig:overview} gives an overview of the proposed framework. Besides a content encoder $G_{enc}$ and a content decoder $G_{dec}$, the proposed method also contains two modules, \emph{i.e.}, a global alignment network and a local alignment network. The purpose of the global alignment network is to replace the source domain style with the target domain style, while the local alignment network is designed to ensure that the translation network focuses on relevant pixels in the image. 
The content encoder $G_{enc}$ and the content decoder $G_{dec}$ of the generator $G$ are applied sequentially to generate an output image, \ie~$\hat{y}{=}G_{dec}(z_x^c){=}G_{dec}(G_{enc}(x))$.
The global alignment network employs a style encoder $f_{enc}$ including shared MLP-Mixer layers \cite{tolstikhin2021mlp} but different fully-connected layers. It extracts style features from source image $x$ and target image $y$. The outputs of the style encoder are denoted as $z_x^s$ and $z_y^s$, respectively.
In detail, $z_x^s$ includes an $N$-dimensional vector of multivariate Gaussian parameters ($\mu_x,\sigma_x$), which represent the style features of the source image $x$. The relationship between the Gaussian parameters $\mu_x,\sigma_x$ and $z_x^s$ is defined as follow:
\begin{equation}
    \mu_x=z_x^s[1:N],\ \ \ \sigma_x=z_x^s[N+1:2N],
    \label{eq:mg}
\end{equation}
Similarly, $z_y^s$ comprises $\mu_y$ and $\sigma_y$.
The global alignment loss $\mathcal{L}_{global}$ aims to make the style features between the two domains as close as possible. The style features of the target image, $\mu_y,\sigma_y$, are also used to provide weights to the IN layer of the content decoder.
The local alignment network integrates a self-attention network $f_{self}$ which provides an attention map $A$.
The source image $x$ and the output image $\hat{y}$, weighted by the attention map $A$ respectively, are provided to the feature extractor $f_{ext}$ supervised by a local alignment loss $\mathcal{L}_{local}$.
 
\subsection{Global Alignment Network}
\label{sec:global}
\par\noindent\textbf{MLP-Mixer-Based Style Encoder.}
Extracting style features plays a vital role in image-to-image translation, which is particularly important in our global alignment network. Different from previous works which use multiple ResBlocks~\cite{kim2020u,Rafique_2021_CVPR} as style encoder, we proposed to employ MLP-Mixer layer~\cite{tolstikhin2021mlp}, which is an effective but conceptually and technically simple alternative, with fully-connected layers (the related experiments in supplementary material further confirms this augment). 
Following~\cite{tolstikhin2021mlp,dosovitskiy2021image}, we divide the input image into a sequence of flattened 2D paths, $x_p{\in}\mathbb{R}^{n\times(P^2\cdot C)}$, and compute per-patch linear embeddings. Then the output of per-patch linear embedding is passed through Mixer layers and fully-connected layers to get the final style feature.
Given the input image $x{\in} \mathbb{R}^{H\times W\times C}$, the full style encoder process is defined as follow:
\begin{align}
    z&=[x_{class};x^1_pE;x^2_pE;\cdots;x^n_pE]+E_{pos},\\
    z'&=z+W_2\phi(W_1 \ LN(x)),\label{eq:mixer1}\\
    Z''&=z'+W_4\phi(W_3 \ LN(x)),\label{eq:mixer2}\\
    Z^s&=FC(Z''),
\end{align}
where $\phi$ is an element-wise nonlinearity~\cite{hendrycks2016gaussian}, $LN$ indicates layer normalization and $W_{1,2,3,4}$ are the weights of the convolution layers. Eq.~\eqref{eq:mixer1} and Eq.~\eqref{eq:mixer2} construct a basic MLP-Mixer layer. We denote the number of MLP-Mixer layers as depth.

\par\noindent\textbf{Improved Adaptive Instance Normalization.}
Inspired by instance normalization (IN)~\cite{ulyanov2016instance}, we adopt an adaptive instance normalization layer in the content encoder to transfer the source style to the target style. The original IN is formulated as:
\begin{equation}
    IN(x)=\gamma\frac{x-\mu(x)}{\sigma(x)}+\kappa,
    \label{eq:in}
\end{equation}
where $\sigma(\cdot)$, $\mu(\cdot)$ denote standard deviation
and mean computed via the spatial dimension within each channel of the input and $\gamma, \kappa$ are learnable scale and shift parameters. According to~\cite{kim2020u}, this method is more suitable for style transfer tasks while failing in image translation tasks that require significant content changes. Recent works~\cite{huang2017arbitrary,zhou2021domain,kim2020u} proposed to fuse the mean and standard deviation of both domains to address this issue. For example, AdaIN~\cite{huang2017arbitrary} replaces $\gamma, \kappa$ in Eq.~\eqref{eq:in} with the mean and standard deviation of the target image,
\begin{equation}
    AdaIN(x)=\sigma(y)\frac{x-\mu(x)}{\sigma(x)}+\mu(y),
\end{equation}

In this work, different from the above methods, we replace the learnable scale and shift parameters in Eq.~\eqref{eq:in} with style features predicted by a style encoder. To sufficiently represent the style features, we compute predictions of $N$ sets $(\mu^i_y,\sigma^i_y)_{i\in 1...N}$. 
Thus, the formula of adaptive instance normalization is updated as:
\begin{equation}
    AdaIN^{new}(x)=\sigma_y\frac{x-\mu(x)}{\sigma(x)}+\mu_y,
    \label{eq:ain}
\end{equation}

\par\noindent\textbf{Style Feature Alignment.}
Having multivariate Gaussian distributions, $\mathcal{N}(\mu_x,\sigma_x^2)$ and $\mathcal{N}(\mu_y,\sigma_y^2)$, from the source domain and the target domain according to Eq.~\eqref{eq:mg}, we adapt a likelihood loss $\mathcal{L}_l$ from~\cite{zhao2017infovae} to force the two multivariate Gaussian distributions to be close to each other. The likelihood loss is defined as:
\begin{equation}
    \mathcal{L}_l=-\frac{1}{2\sigma^2_x}\sum_{i=1}^N(s_y^i-\mu_x)^2,
\end{equation}
where $s_y$ are $N$ sample from multivariate Gaussian distribution $\mathcal{N}(\mu_y,\sigma_y^2)$. The number of samples is equal to the dimensions of predicted Gaussian distributions in any domain. We also employ a regularization loss $\mathcal{L}_r$ to encourage the model to predict various Gaussian distributions and to learn meaningful latent representations according to InfoVAE~\cite{zhao2017infovae}. The regularization loss $\mathcal{L}_r$ is calculated by:
\begin{equation}
    \mathcal{L}_r=-D_{KL}(\mathcal{N}(0,1),\mathcal{N}(\mu_x,\sigma_x^2)),
\end{equation}
where $D_{KL}$ represents Kullback–Leibler divergence and $\mathcal{N}(0,1)$ represents a unit Gaussian. Therefore, the whole global alignment loss is as follows:
\begin{equation}
    \mathcal{L}_{global}=\lambda_l \mathcal{L}_l+\lambda_r \mathcal{L}_r,
    \label{eq:global}
\end{equation}
where both $\lambda_l$ and $\lambda_r$ are set to 1  following~\cite{zhao2017infovae}. 

\subsection{Local Alignment Network}
\label{sec:local}
The global alignment network ensures accurate style transfer. However, for image-to-image translation tasks with significant content changes, the network will easily focus on unimportant pixels, as discussed in Section~\ref{sec:ab}.
To address this problem, we propose a novel local alignment network supervised by a pixel-wise spatial-correlative loss.
We consider DeiT-S-p8~\cite{caron2021emerging} as our self-attention network $f_{self}$ and compute the attention map $A$ as $A{=}f_{self}(x)$ and subsequently a spatially-correlative map~\cite{zheng2021spatially},
\begin{equation}
    S_x=f_{ext}(A\cdot x)^Tf_{ext}(A\cdot x)_*,
\end{equation}
where $(\cdot)_*$ indicates corresponding features of image $x$ in a path of points. {In the same way, replacing $x$ with $\hat{y}$ in the above equation, we can get the spatially-correlative map of generated image $S_{\hat{y}}$.}
Additionally, we define the pixel-wise spatial-correlative loss as follow:
\begin{equation}
    \mathcal{L}_{local}=||1-\cos(S_x,S_{\hat{y}})||,
    \label{eq:local}
\end{equation}
This loss helps the generator to avoid content destruction in the unrelevant image part. In fact, it evaluates every pixel importance by the spatially-correlative map and supports the spatial similarity to be consistent at all points.

\subsection{Overall Optimization Objective}
\label{sec:opt}
Our framework also integrates a discriminator, as shown in Figure~\ref{fig:overview}. 
Hence, besides the global alignment loss and local alignment loss, our framework also employs a generative adversarial loss. This adversarial loss $\mathcal{L}_{GAN} {=} \mathcal{L}_{GAN}^D {+} \mathcal{L}_{GAN}^G$ can be formulated as follows,
\begin{align}
    \mathcal{L}_{GAN}^D &=-\mathbb{E}[\log D(y)]-\mathbb{E}[\log (1-D(\hat{y}))],\\
    \mathcal{L}_{GAN}^G &=\mathbb{E}[\log (1-D(\hat{y}))],
\end{align}

The whole optimization objective of the proposed GLA-Net can be expressed as:
\begin{equation}
    \mathcal{L}=\mathcal{L}_{GAN}+\lambda_{global} \mathcal{L}_{global}+\lambda_{local} \mathcal{L}_{local},
\end{equation}
where $\lambda_{global}$ and $\lambda_{local}$ are hyper-parameters are controlling the importance of the corresponding loss term, which are empirically set to 1 and 10 in all our experiments, respectively. The detailed parameter selection process is discussed in the supplementary material.
\begin{figure*}[!t] 
\centering
\includegraphics[width=1\textwidth]{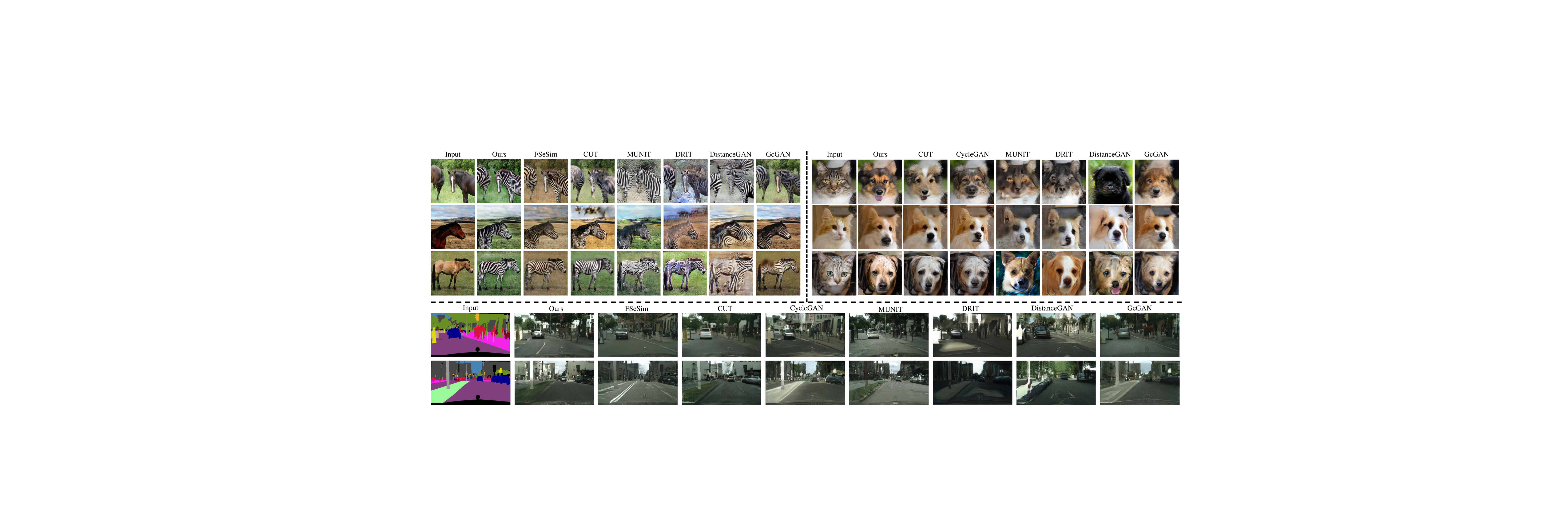}
\caption{Qualitative results with the several challenging methods, \ie FSeSim~\cite{zheng2021spatially}, CUT~\cite{park2020contrastive}, CycleGAN~\cite{zhu2017unpaired}, MUNIT~\cite{huang2018multimodal}, DRIT~\cite{lee2018diverse}, DistanceGAN~\cite{benaim2017one}, GcGAN~\cite{fu2019geometry}, on three benchmark datasets of single-modal image translation. 
}
\label{fig:vis}
\vspace{-0.4cm}
\end{figure*}

\section{Experiments}
\subsection{Experiment Setups}
\noindent \textbf{Datasets.} We tested our method on various image-to-image translation tasks, including single-modal (Cityscapes~\cite{cordts2016cityscapes}, Cat$\to$Dog~\cite{choi2020stargan}, Horse$\to$Zebra~\cite{zhu2017unpaired}) and multi-modal translation (Winter$\to$Summer~\cite{zhu2017unpaired}, Night$\to$Day~\cite{isola2017image}) problems.
The Cityscapes dataset contains thousands of finely annotated images taken in street scenes with pixel-level annotations.
Cat$\to$Dog contains 5,000 training and 500 validation images from AFHQ dataset~\cite{choi2020stargan}.
The Horse$\to$Zebra dataset consists of about 2,500 images of horse and zebra in different scenes.
Winter$\to$Summer~\cite{zhu2017unpaired} is downloaded using Flickr API with the tag ``yosemite'' and the ``datetaken'' field, including 1,273 summer images and 854 winter images.
Night$\to$Day, trained on~\cite{laffont2014transient}, includes 17,823 training images extracted from 91 webcams.
For all the datasets, we resize the images to the same resolution of 256$\times$256 pixels. 

\noindent \textbf{Implementation Details.}
We choose FSeSim~\cite{zheng2021spatially} as the baseline architecture and replace the spatially-correlative loss with our local alignment loss.
Specifically, we use the UNet-based content encoder and decoder with AdaIN~\cite{karras2019style} in the global alignment network. We also adopt MLP-Mixer with depth 1 as our style encoder in the global alignment network. The ImageNet-pretrained VGG16~\cite{SimonyanZ14a} is chosen as our feature extractor, where we use feature from layers conv2d$\_\{4,7,9\}$.
For all the experiments, we use the Adam solver~\cite{KingmaB14} with a batch size of 1. All
networks were trained from scratch with a learning rate of $1{\times} 10^{-4}$. The dimension of the predicted multivariate Gaussian distribution $N$ is set to 32 for all experiments.
The training lasts 400 epochs in total. 
We apply PyTorch to implement our framework, and we perform all experiments on an NVIDIA Titan XP~GPU.

\begin{table*}[t]
\centering
\caption{Quantitative comparison on single-modal image translation. $*$ means results come from our re-implementation. KID means Kernel Inception Distance $\times$100.}
\label{tab:sc}
\resizebox{0.65\linewidth}{!}{%
\begin{tabular}{lccccccccc}
\toprule
\multirow{2}{*}{Method} & \multirow{2}{*}{One-Sided} & \multicolumn{4}{c}{Cityscapes} & \multicolumn{2}{c}{Cat $\to$ Dog}   & \multicolumn{2}{c}{Horse $\to$ Zebra} \\
 \cmidrule(lr){3-6}  \cmidrule(lr){7-8} \cmidrule(lr){9-10} & & mAP $\uparrow$   & pAcc $\uparrow$  & cAcc $\uparrow$  & FID $\downarrow$    & FID $\downarrow$  & KID $\downarrow$   & FID $\downarrow$  & KID $\downarrow$   \\
\midrule
CycleGAN~\cite{zhu2017unpaired} & {\color{red} \xmark} & 20.4 & 55.9 & 25.4 & 76.3 & 85.9 & 6.93 & 77.2 & 3.24 \\
MUNIT~\cite{huang2018multimodal} & {\color{red} \xmark} & 16.9 & 56.5 & 22.5 & 91.4 & 104.4 & 2.42 & 133.8 & 6.92\\
DRIT~\cite{lee2018diverse} & {\color{red} \xmark}  & 17.0 & 58.7 & 22.2 & 155.3 & 123.4 & 4.57 & 140.0 & 7.40\\
NICE-GAN~\cite{chen2020reusing} & {\color{red} \xmark} & - & - & - & -  & 48.8 & 1.58 & 65.9 & 2.09\\
AttentionGAN~\cite{tang2021attentiongan} & {\color{red} \xmark} & - & - & - & - & - & - & 68.6 & 2.03\\
U-GAT-IT~\cite{kim2020u} & {\color{red} \xmark} & - & - & - & -  & -  & 7.07 & -  & 7.06 \\
CWT-GAN~\cite{lai2021unsupervised}& {\color{red} \xmark} & - & - & - & - & \textbf{46.3} & - & 85.4 & - \\
IrwGAN~\cite{xie2021unaligned}& {\color{red} \xmark} & - &	- &	- &	- &	61.0 & \textbf{ 2.07} & 79.4 & 1.83  \\
\midrule
DistanceGAN~\cite{benaim2017one}& {\color{green} \cmark} & 8.4 & 42.2 & 12.6 & 81.8 & 155.3 & - & 72.0 & -\\
AGGAN~\cite{mejjati2018unsupervised}& {\color{green} \cmark} & - & - & - & - & - & 9.84 & - & 6.93\\
GcGAN~\cite{fu2019geometry}& {\color{green} \cmark} & 21.2 & 63.2 & 26.6 & 105.2 & 96.6 & - & 86.7 & -\\
CUT~\cite{park2020contrastive}& {\color{green} \cmark} & \textbf{24.7}  & 68.8  & 30.7 & 56.4 & 76.2 & - & 45.5 & - \\
FSeSim~\cite{zheng2021spatially}& {\color{green} \cmark} & -  & 69.4 & -  & 53.6 & 78.9$^*$ & 3.72$^*$ & 40.4 & -\\
GLA-Net (Ours) & {\color{green} \cmark} & 23.5  & \textbf{76.2}  & \textbf{31.8}  & \textbf{51.8} & {66.6} & 2.94 & \textbf{37.7} & \textbf{0.87}\\
\bottomrule
\end{tabular}}
\vspace{-0.4cm}
\end{table*}

\noindent \textbf{Evaluation Protocols.} Following the evaluation protocols from~\cite{heusel2017gans,zhu2017unpaired,park2020contrastive,zheng2021spatially}, we choose a few metrics to assess the visual quality and measure the domain distance. 
For the first, we utilize the Fr\'echet Inception Distance (FID)~\cite{heusel2017gans}, which empirically estimates the distribution of target and generated images in a deep network space and computes the divergence between them. In this paper, it is used for both single- and multi-modal image translation tasks.
We also use Kernel Inception Distance (KID)~\cite{binkowski2018demystifying}, which is the squared maximum mean discrepancy between Inception representations, to compare with methods only reporting KID results.
To evaluate the Cityscapes task, we feed the generated images to the pretrained semantic segmentation network DRN~\cite{yu2017dilated}. Then the outputs of DRN are used to compute the mean average precision (mAP), pixel-wise accuracy (pAcc), and average class accuracy (cAcc), as done in~\cite{zhang2016colorful,isola2017image,park2020contrastive,park2019semantic}.
Meanwhile, we also use the average LPIPS distance~\cite{zhang2018unreasonable} to evaluate the multi-modal image translation, which measures the distance between two images in a feature domain and correlates well with human perception.
Finally, we consider density and coverage (D\&C)~\cite{naeem2020reliable}, which uses density and coverage to measure similarity between the generated manifold and the target manifold. 

\subsection{Single-Modal Unpaired Image Translation}

We compare our method with state-of-the-art methods,~\ie~U-GAT-IT~\cite{kim2020u}, CWT-GAN~\cite{lai2021unsupervised}, IrwGAN~\cite{xie2021unaligned}, DistanceGAN~\cite{benaim2017one}, GcGAN~\cite{fu2019geometry}, CUT~\cite{park2020contrastive}, and FSeSim~\cite{zheng2021spatially} on single-modal image translation. According to whether the methods are one-sided framework, Table~\ref{tab:sc} is divided into upper (double-sided framework) and lower (one-sided framework) parts for a fair comparison. We pick up three popular datasets on single-modal image translation, \ie~cityscapes, cat$\to$dog, and horse$\to$zebra, all of 
which belong to the task requiring style transfer and large content changes. Unsurprisingly, we find the methods which solely focus on style transfer like U-GAT-IT~\cite{kim2020u} and the attention-based networks, such as AGGAN~\cite{mejjati2018unsupervised} and AttentionGAN~\cite{tang2021attentiongan}, obtain unsatisfactory performance on these three tasks. In contrast, 
since our approach comprises both a style transform network and an attention-based network, it achieves the best results considering most of the metrics in Table~\ref{tab:sc}.
Moreover, comparing our method with other one-sided methods,~\ie~CUT~\cite{park2020contrastive} and FSeSim~\cite{zheng2021spatially}, our method significantly outperforms these methods. 

Qualitative results with several challenging methods on single-modal image translation are shown in Figure~\ref{fig:vis}.
We find that the images generated by our method have higher quality and sharper details
than those obtained with previous approaches. This confirms the fact that since our work seamlessly integrates two components, it ensures higher flexibility in style and content modification. Our network better focuses and changes the most important parts of the image while preserving the background information. 

\subsection{Multi-Modal Unpaired Image Translation}
\begin{figure}[!t]
\centering
\subfloat[Input]{
    \begin{minipage}{0.24\linewidth}
        \centering
        \includegraphics[width=0.993\textwidth,height=0.5in]{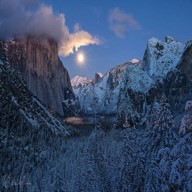}\\
        \includegraphics[width=0.993\textwidth,height=0.5in]{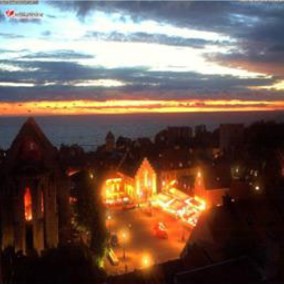}\\
        \includegraphics[width=0.993\textwidth,height=0.5in]{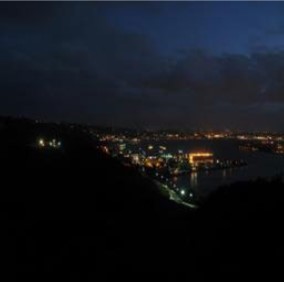}\\  
        \includegraphics[width=0.993\textwidth,height=0.5in]{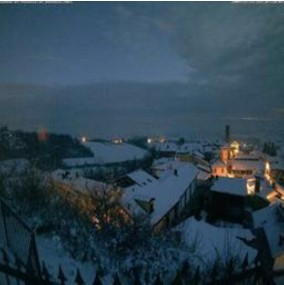}\\
        \includegraphics[width=0.993\textwidth,height=0.5in]{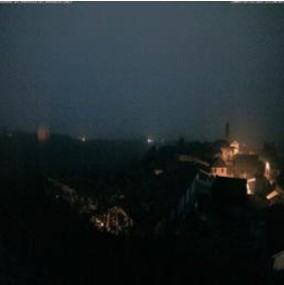}\\       
    \end{minipage}%
}%
\subfloat[Ours]{
    \begin{minipage}{0.24\linewidth}
        \centering
        \includegraphics[width=0.993\textwidth,height=0.5in]{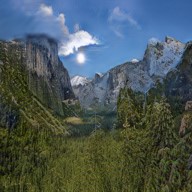}\\
        \includegraphics[width=0.993\textwidth,height=0.5in]{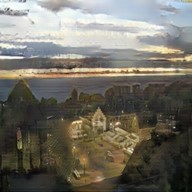}\\
        \includegraphics[width=0.993\textwidth,height=0.5in]{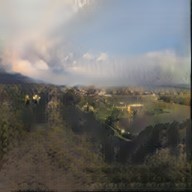}\\  
        \includegraphics[width=0.993\textwidth,height=0.5in]{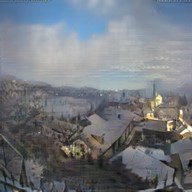}\\
        \includegraphics[width=0.993\textwidth,height=0.5in]{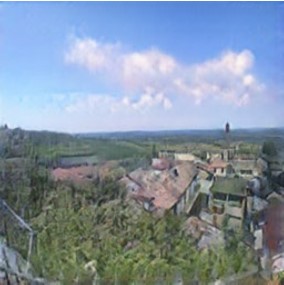}\\   
    \end{minipage}%
}%
\subfloat[FSeSim~\cite{zheng2021spatially}]{
    \begin{minipage}{0.24\linewidth}
        \centering
        \includegraphics[width=0.993\textwidth,height=0.5in]{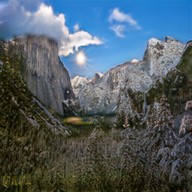}\\
        \includegraphics[width=0.993\textwidth,height=0.5in]{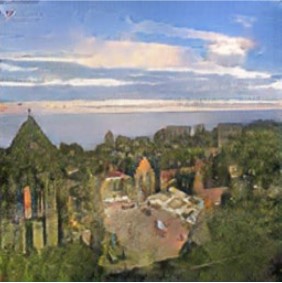}\\
        \includegraphics[width=0.993\textwidth,height=0.5in]{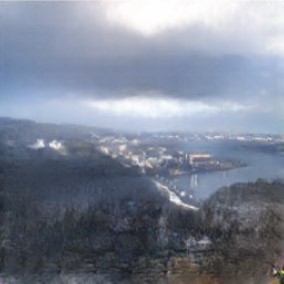}\\  
        \includegraphics[width=0.993\textwidth,height=0.5in]{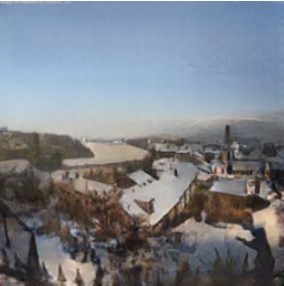}\\
        \includegraphics[width=0.993\textwidth,height=0.5in]{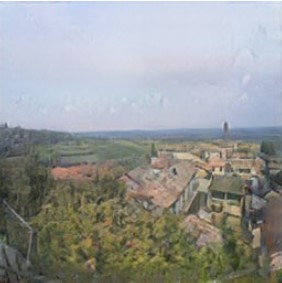}\\   
    \end{minipage}%
}%
\subfloat[MUNIT~\cite{huang2018multimodal}]{
    \begin{minipage}{0.24\linewidth}
        \centering
        \includegraphics[width=0.993\textwidth,height=0.5in]{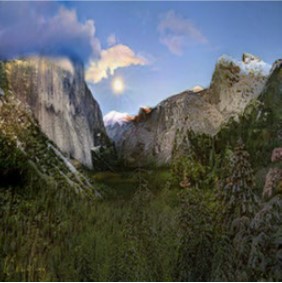}\\
        \includegraphics[width=0.993\textwidth,height=0.5in]{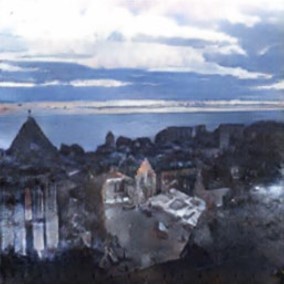}\\
        \includegraphics[width=0.993\textwidth,height=0.5in]{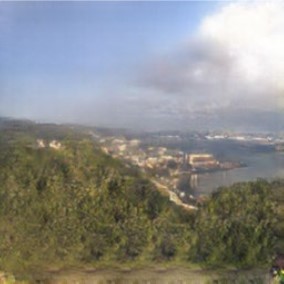}\\  
        \includegraphics[width=0.993\textwidth,height=0.5in]{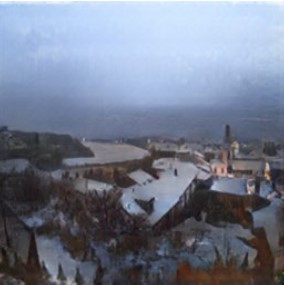}\\
        \includegraphics[width=0.993\textwidth,height=0.5in]{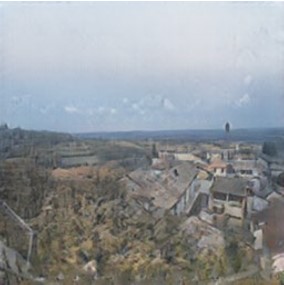}\\  
    \end{minipage}%
}%
\centering
\caption{Qualitative results on two datasets, \ie~Winter$\to$Summer and Day$\to$Night, of multi-modal image translation. }
\label{fig:vis_multi}
\vspace{-0.4cm}
\end{figure}

To verify the versatility of our algorithm, we also provide a comparison among our method and other state-of-the-art methods, \ie~NICE-GAN~\cite{chen2020reusing}, FSeSim~\cite{zheng2021spatially} and CWT-GAN~\cite{lai2021unsupervised}, on multi-modal unpaired image translation tasks. The chosen benchmarks include Winter$\to$Summer dataset and Night$\to$Day datasets, which are regarded as typical style transfer tasks. According to the results in Table~\ref{tab:mm}, our method significantly surpasses FSeSim~\cite{zheng2021spatially}. In detail, in terms of FID, there is an 18.6 and 43.5 improvement between our method and FSeSim on both datasets, respectively. Moreover, our method achieves better results than the recent SOTA, CWT-GAN~\cite{lai2021unsupervised} and NICE-GAN~\cite{chen2020reusing}. Qualitative results on multi-modal image translation are shown in Figure~\ref{fig:vis_multi}. 
What is more, in terms of D\&C, our method reaches the new SOTA leaving a considerable margin with other methods in the Winter$\to$Summer task while our method also gets an acceptable result.
Similar to single-modal image translation, the results generated by our method are more photo-realistic than those obtained by previous methods.

\begin{table}[t] \small
\centering
\caption{Qualitative comparison on multi-modal image translation.}
\label{tab:mm}
\resizebox{0.98\linewidth}{!}{%
\begin{tabular}{lcccc}
\toprule
\multirow{2}{*}{Method} & \multicolumn{2}{c}{Winter $\to$ Summer} & \multicolumn{2}{c}{Night $\to$ Day} \\
\cmidrule(lr){2-3} \cmidrule(lr){4-5} & FID $\downarrow$ & D \& C $\uparrow$   & FID $\downarrow$  & D \& C $\uparrow$   \\
\midrule
BicycleGAN~\cite{wu2019transgaga} & 99.2   & 0.536 / 0.667  & 290.9  & \textbf{0.375} / 0.515  \\
MUNIT~\cite{huang2018multimodal} & 97.4   & 0.439 / 0.707  & 267.1  & 0.271 / 0.548  \\
DRIT++~\cite{lee2020drit++}  & 93.1   & 0.494 / 0.753  & 258.5  & 0.298 / 0.631  \\
FSeSim~\cite{zheng2021spatially}  & 90.5   & 0.501 / 0.779  & 234.3  & 0.332 / \textbf{0.638}  \\
CWT-GAN~\cite{lai2021unsupervised} & 77.0 & - & - & - \\
NICE-GAN~\cite{chen2020reusing} & 76.4& - & - & - \\
GLA-Net (Ours) & \textbf{71.9}  & \textbf{0.879 / 0.894}  & \textbf{190.8} & {0.301 / 0.629} \\
\bottomrule
\end{tabular}
}
\vspace{-0.4cm}
\end{table}

\subsection{Ablation Study}
\label{sec:ab}
To demonstrate the effectiveness of different components of the proposed GLA-Net, we train several model variants and test them on both single- and multi-modal unpaired image translation tasks. The results are reported in Table~\ref{tab:ab}.
We choose FSeSim~\cite{zheng2021spatially} as our baseline, and reproduced results are reported in the first row of Table~\ref{tab:ab}.
According to results in Table~\ref{tab:ab}, adding a global alignment network or a local alignment network significantly improves performance according to all metrics. It proves the effectiveness of both sub-networks. 

\begin{figure}[!t]
\centering
\subfloat[Input]{
    \begin{minipage}{0.24\linewidth}
        \centering
        \includegraphics[width=0.993\textwidth,height=0.5in]{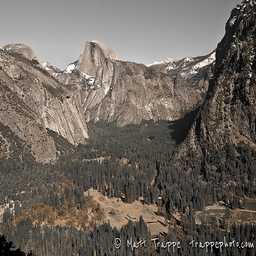}\\
        \includegraphics[width=0.993\textwidth,height=0.5in]{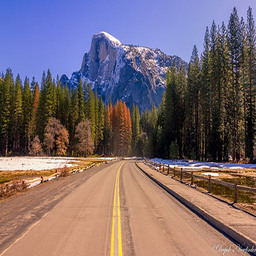}\\
        \includegraphics[width=0.993\textwidth,height=0.5in]{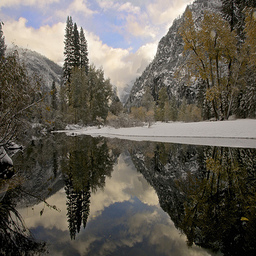}\\  
        \includegraphics[width=0.993\textwidth,height=0.5in]{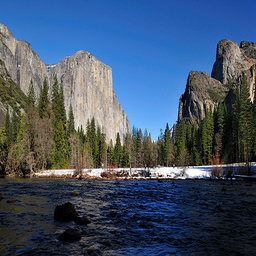}\\
        \includegraphics[width=0.993\textwidth,height=0.5in]{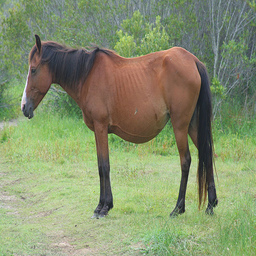}\\
        \includegraphics[width=0.993\textwidth,height=0.5in]{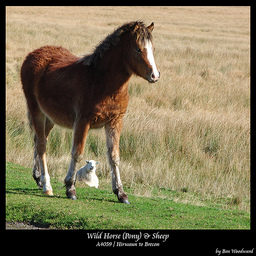}\\        
        
    \end{minipage}%
}%
\subfloat[Attention]{
    \begin{minipage}{0.24\linewidth}
        \centering
        \includegraphics[width=0.993\textwidth,height=0.5in]{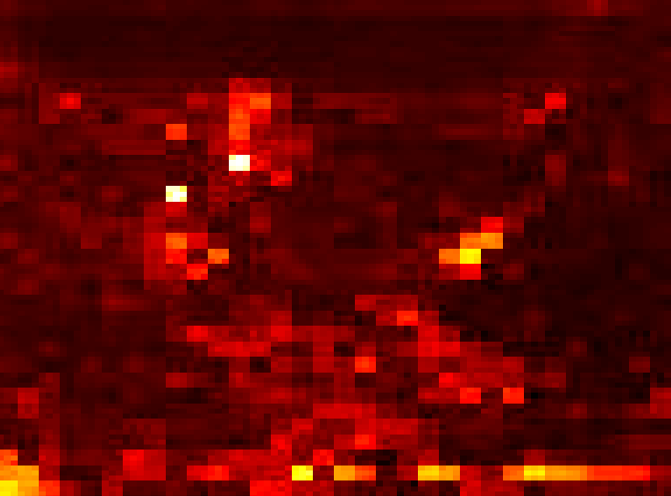}\\
        \includegraphics[width=0.993\textwidth,height=0.5in]{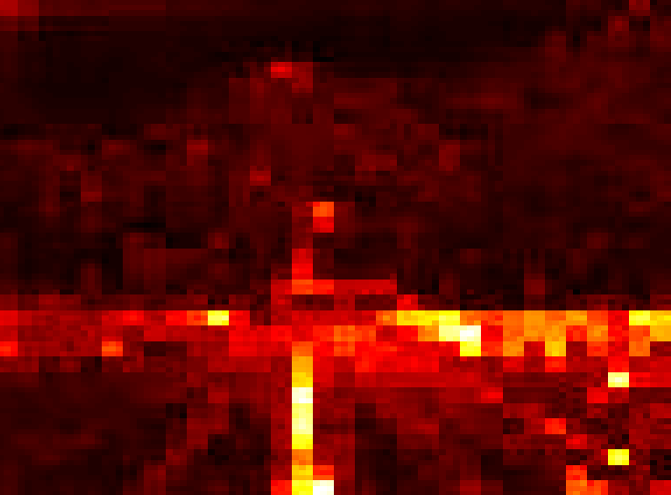}\\
        \includegraphics[width=0.993\textwidth,height=0.5in]{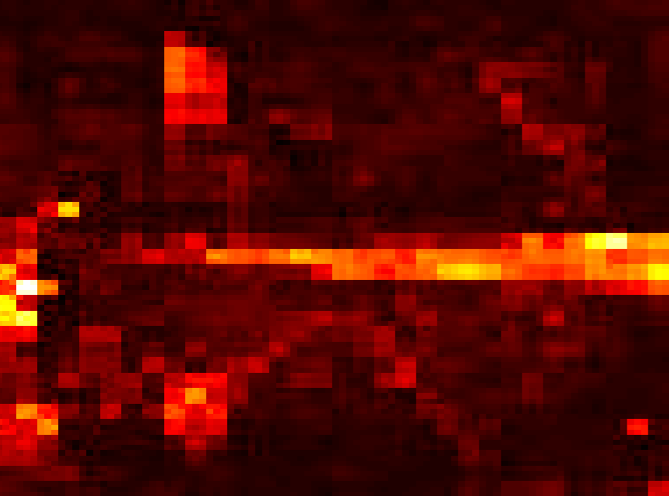}\\  
        \includegraphics[width=0.993\textwidth,height=0.5in]{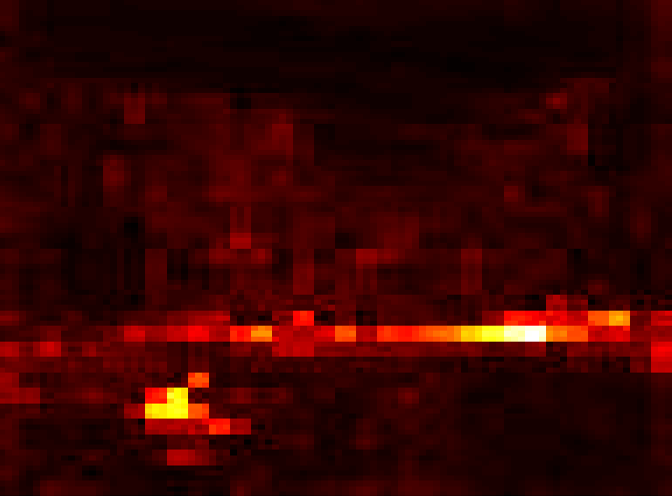}\\
        \includegraphics[width=0.993\textwidth,height=0.5in]{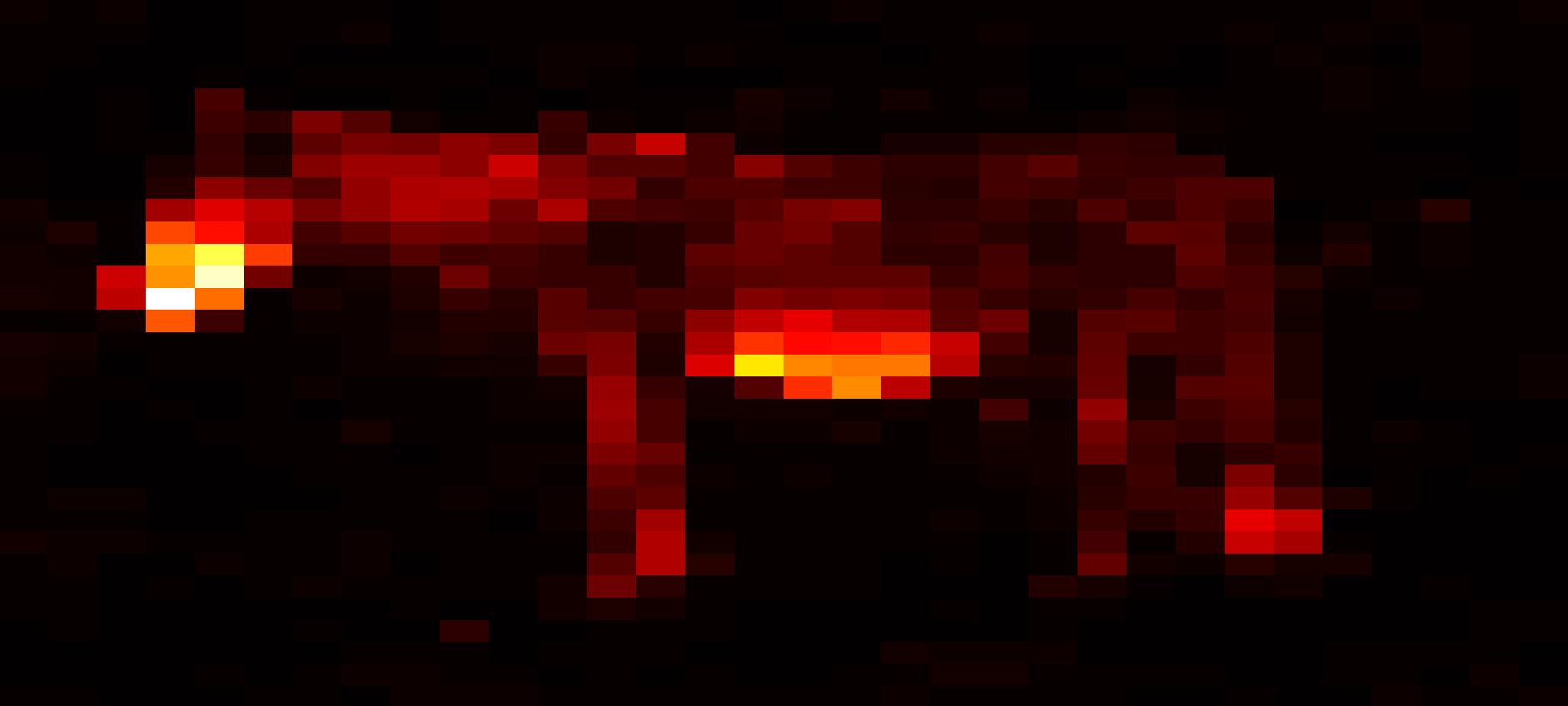}\\
        \includegraphics[width=0.993\textwidth,height=0.5in]{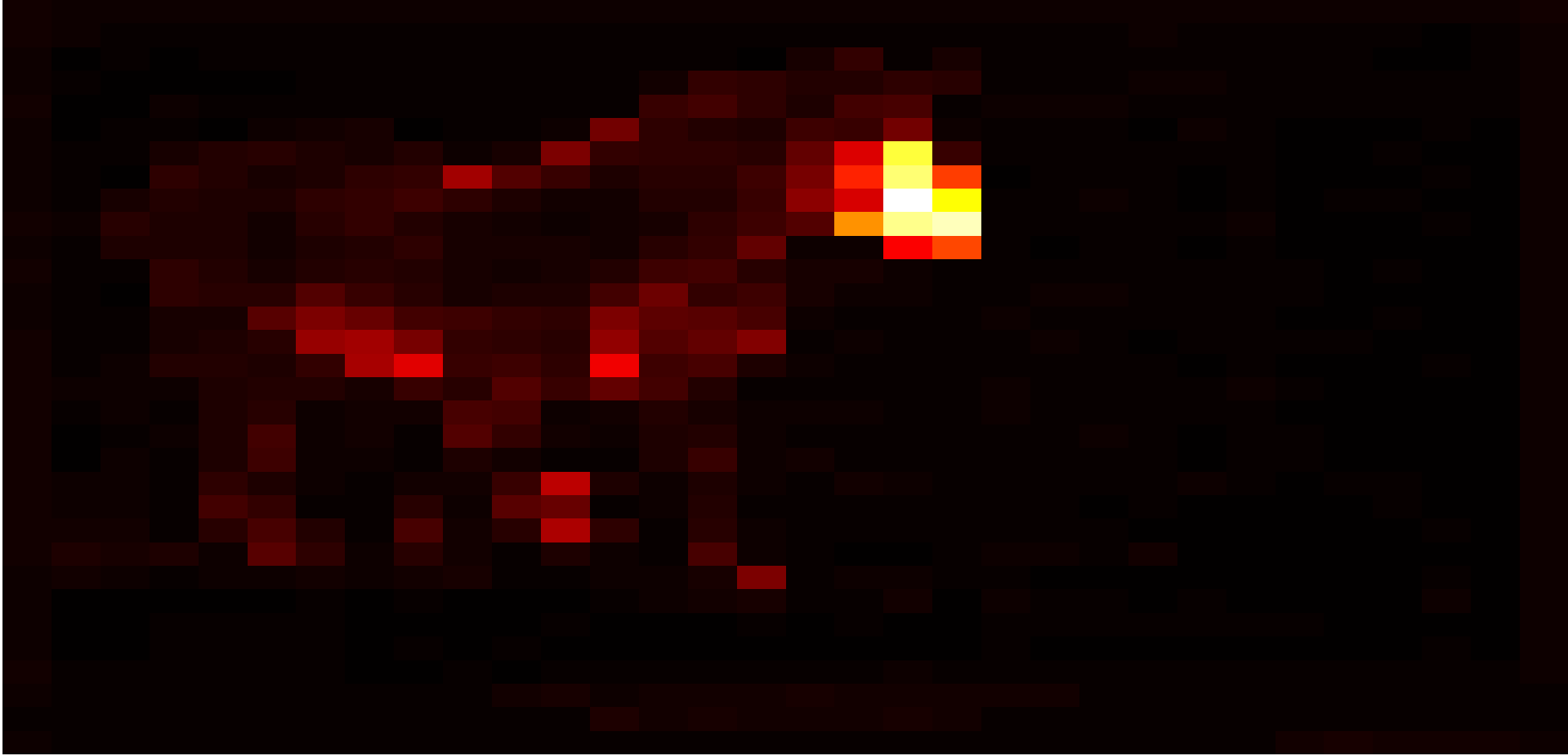}\\        
    \end{minipage}%
}%
\subfloat[Baseline]{
    \begin{minipage}{0.24\linewidth}
        \centering
        \includegraphics[width=0.993\textwidth,height=0.5in]{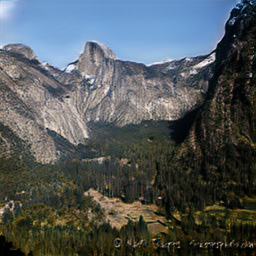}\\
        \includegraphics[width=0.993\textwidth,height=0.5in]{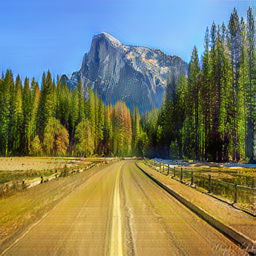}\\
        \includegraphics[width=0.993\textwidth,height=0.5in]{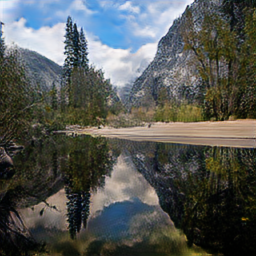}\\  
        \includegraphics[width=0.993\textwidth,height=0.5in]{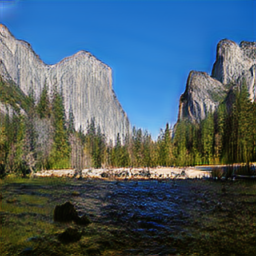}\\
        \includegraphics[width=0.993\textwidth,height=0.5in]{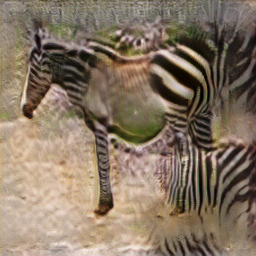}\\
        \includegraphics[width=0.993\textwidth,height=0.5in]{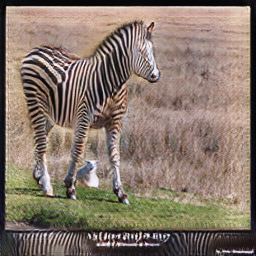}\\        
    \end{minipage}%
}%
\subfloat[w/ $\mathcal{L}_{local}$]{
    \begin{minipage}{0.24\linewidth}
        \centering
        \includegraphics[width=0.993\textwidth,height=0.5in]{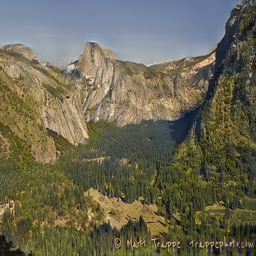}\\
        \includegraphics[width=0.993\textwidth,height=0.5in]{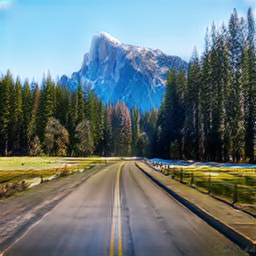}\\
        \includegraphics[width=0.993\textwidth,height=0.5in]{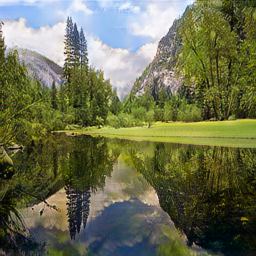}\\  
        \includegraphics[width=0.993\textwidth,height=0.5in]{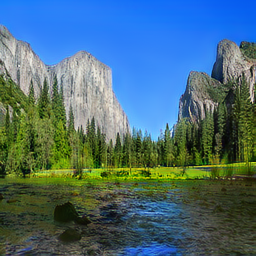}\\
        \includegraphics[width=0.993\textwidth,height=0.5in]{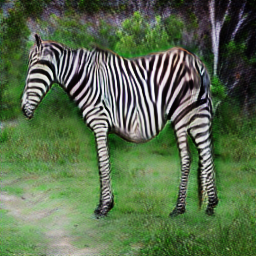}\\     
        \includegraphics[width=0.993\textwidth,height=0.5in]{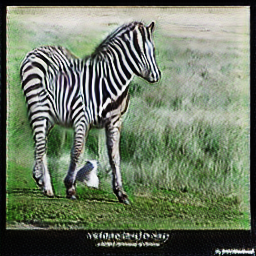}\\         
    \end{minipage}%
}%
\centering
\caption{Visualisation of generated attention maps on two benchmark datasets, \ie~Winter$\to$Summer and Horse $\to$ Zebra.}
\label{fig:vis_att}
\vspace{-0.4cm}
\end{figure}

We provide some visualization results of our attention maps in Figure~\ref{fig:vis_att}. To demonstrate the impact of the attention, the input image, the output of the baseline method, and the output of baseline with the local alignment network are shown in Figure~\ref{fig:vis_att}. 
The pixels whose color is closer to yellow are regarded as more important by the network. Conversely, darker pixels are considered less relevant. 
The figure clearly shows that when adding the attention map into the baseline model, the visual quality of important pixels gets significant improvements. 
It proves that the local alignment network effectively helps the generator pay more attention to important pixels during translation. 

\begin{figure}[t]
    \centering
    \includegraphics[width=0.95\linewidth]{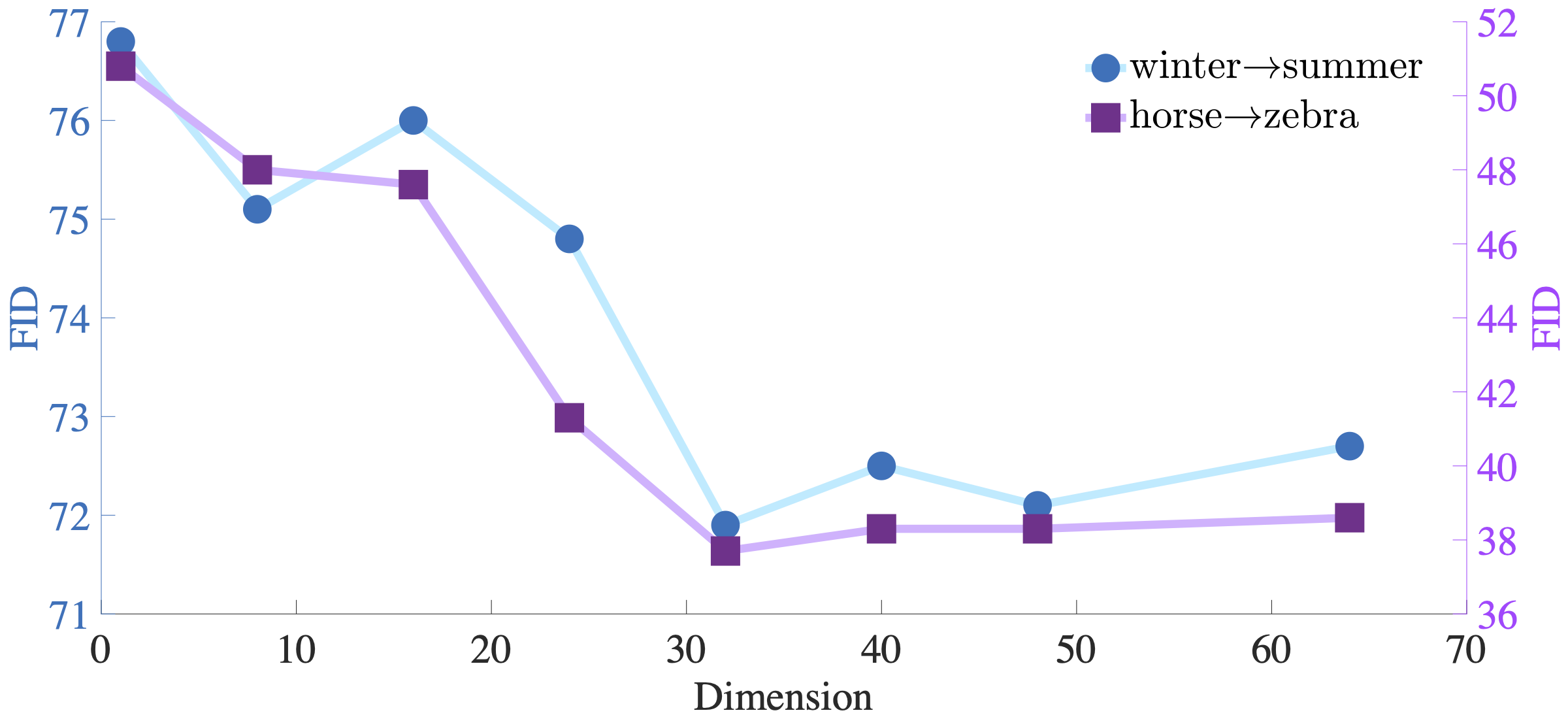}
    \caption{Ablation study of the dimensions of predicted multivariate Gaussian distribution.
 }
    \label{fig:a}
    \vspace{-0.4cm}
\end{figure}

\begin{figure*}[!t]
\centering
\subfloat[Input Image]{
    \begin{minipage}{0.15\linewidth}
        \centering
        \includegraphics[width=0.993\textwidth,height=0.7in]{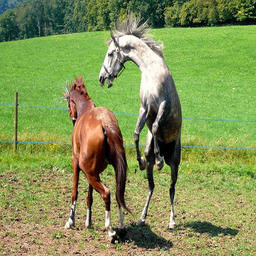}\\
        \includegraphics[width=0.993\textwidth,height=0.7in]{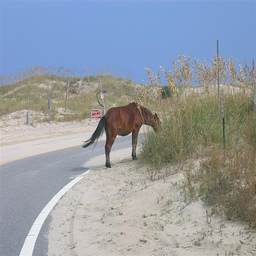}\\
        \includegraphics[width=0.993\textwidth,height=0.7in]{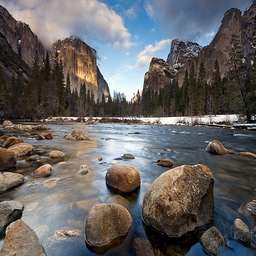}\\  
        \includegraphics[width=0.993\textwidth,height=0.7in]{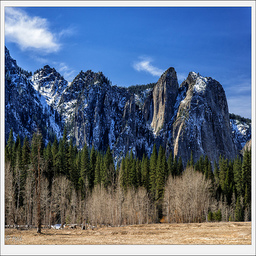}\\
    \end{minipage}%
}%
\subfloat[Baseline]{
    \begin{minipage}{0.15\linewidth}
        \centering
        \includegraphics[width=0.993\textwidth,height=0.7in]{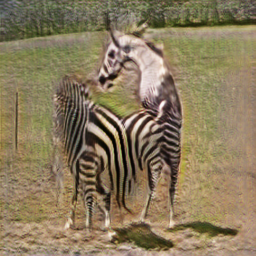}\\
        \includegraphics[width=0.993\textwidth,height=0.7in]{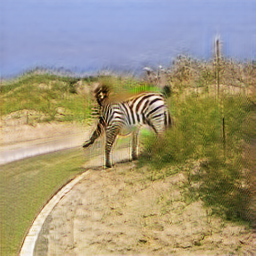}\\
        \includegraphics[width=0.993\textwidth,height=0.7in]{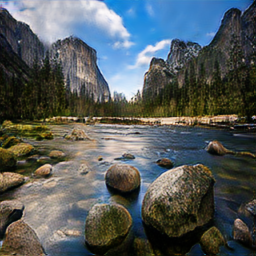}\\  
        \includegraphics[width=0.993\textwidth,height=0.7in]{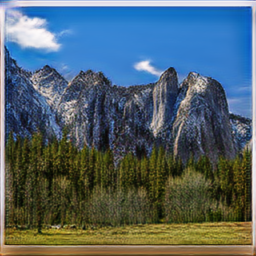}\\
    \end{minipage}%
}%
\subfloat[w/ AdaIN]{
    \begin{minipage}{0.15\linewidth}
        \centering
        \includegraphics[width=0.993\textwidth,height=0.7in]{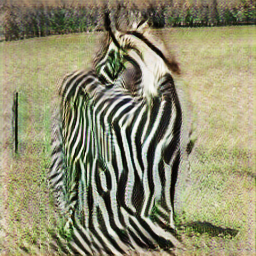}\\
        \includegraphics[width=0.993\textwidth,height=0.7in]{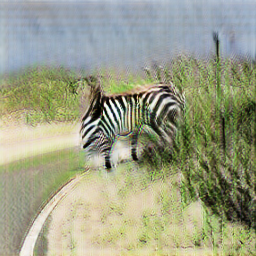}\\
        \includegraphics[width=0.993\textwidth,height=0.7in]{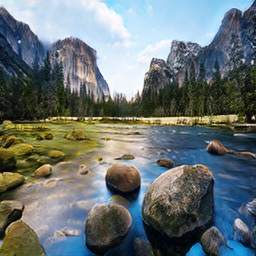}\\
        \includegraphics[width=0.993\textwidth,height=0.7in]{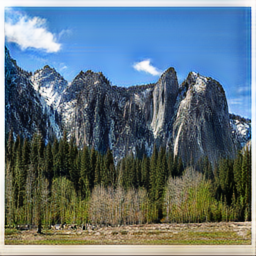}\\
    \end{minipage}%
}%
\subfloat[w/ Global Alig.]{
    \begin{minipage}{0.15\linewidth}
        \centering
        \includegraphics[width=0.993\textwidth,height=0.7in]{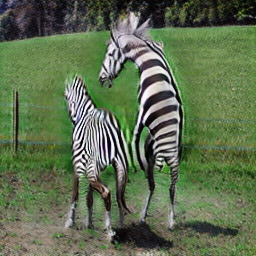}\\
        \includegraphics[width=0.993\textwidth,height=0.7in]{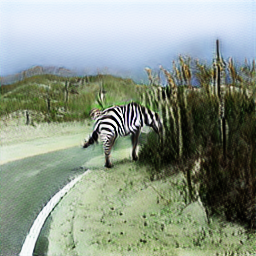}\\
        \includegraphics[width=0.993\textwidth,height=0.7in]{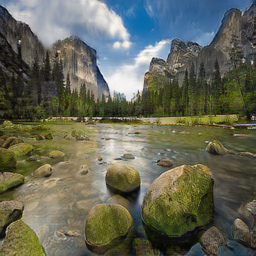}\\  
        \includegraphics[width=0.993\textwidth,height=0.7in]{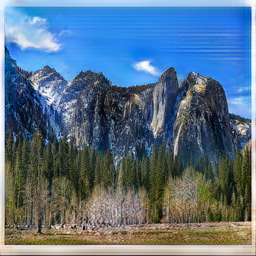}\\
    \end{minipage}%
}%
\subfloat[w/ Local Alig.]{
    \begin{minipage}{0.15\linewidth}
        \centering
        \includegraphics[width=0.993\textwidth,height=0.7in]{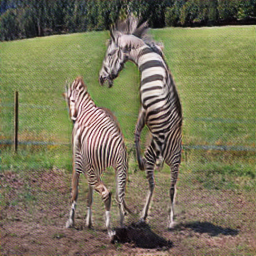}\\
        \includegraphics[width=0.993\textwidth,height=0.7in]{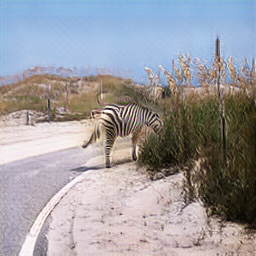}\\
        \includegraphics[width=0.993\textwidth,height=0.7in]{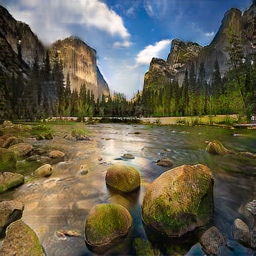}\\  
        \includegraphics[width=0.993\textwidth,height=0.7in]{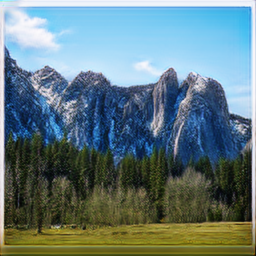}\\
    \end{minipage}%
}%
\subfloat[GLA-Net (Our Full)]{
    \begin{minipage}{0.15\linewidth}
        \centering
        \includegraphics[width=0.993\textwidth,height=0.7in]{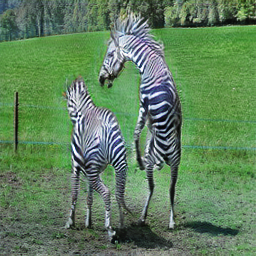}\\
        \includegraphics[width=0.993\textwidth,height=0.7in]{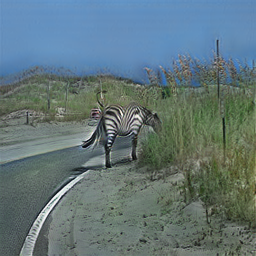}\\
        \includegraphics[width=0.993\textwidth,height=0.7in]{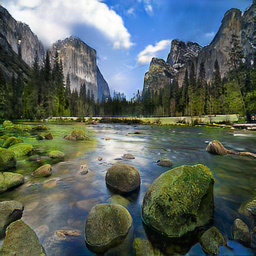}\\ 
        \includegraphics[width=0.993\textwidth,height=0.7in]{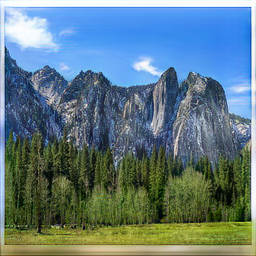}\\
    \end{minipage}%
}%
\centering
\caption{Visualisation results of different methods in Table~\ref{tab:ab}. 
}
\label{fig:vis_ab}
\vspace{-0.4cm}
\end{figure*}

\begin{table*}[h]
\centering
\caption{Ablation study of our global and local alignment method on both single- and multi-modal unpaired image translation. $AdaIN^{new}$, $\mathcal{L}_{global}$ and $\mathcal{L}_{local}$ is defined by Eq.~\eqref{eq:ain}, \eqref{eq:global}, and \eqref{eq:local}, respectively.}
\label{tab:ab}
\resizebox{0.85\linewidth}{!}{%
\begin{tabular}{l|ccc|cccccc}
\toprule
\multirow{2}{*}{Method} & \multicolumn{1}{c}{\multirow{2}{*}{$AdaIN^{new}$}} & \multicolumn{1}{c}{\multirow{2}{*}{$\mathcal{L}_{global}$}} & \multicolumn{1}{c|}{\multirow{2}{*}{$\mathcal{L}_{local}$}} & \multicolumn{3}{c}{Horse $\to$ Zebra} & \multicolumn{3}{c}{Winter $\to$ Summe} \\  \cmidrule(lr){5-7} \cmidrule(lr){8-10}
 & \multicolumn{1}{c}{} & \multicolumn{1}{c}{} & \multicolumn{1}{c|}{} & \multicolumn{1}{c}{LPIPS $\downarrow$} & \multicolumn{1}{c}{FID $\downarrow$} & \multicolumn{1}{c}{D \& C $\uparrow$} & \multicolumn{1}{c}{LPIPS $\downarrow$} & \multicolumn{1}{c}{FID $\downarrow$} & \multicolumn{1}{c}{D \& C $\uparrow$} \\
\midrule
Baseline &  &   &  &  0.743& 46.5 & 0.825/ 0.800   &  0.792 & 89.1 & 0.727/0.739 \\ \hline
w/ AdaIN & $\checkmark$ &  &  & 0.767  & 51.7 & 0.819/ 0.794 & 0.758  & 79.0 & 0.767/0.839 \\
w/ Global Alignment & $\checkmark$ & $\checkmark$ &  & 0.722 & 39.7  & 0.966/0.936 & 0.761 & 74.8 & 0.856/0.841 \\
\hline
w/ Local Alignment &  & & $\checkmark$  & 0.732 & 40.6 & 0.953/ 0.924 & 0.755 & 78.4 & 0.869/0.846  \\
GLA-Net (Our Full)  & $\checkmark$ & $\checkmark$ & $\checkmark$ & \textbf{0.714} & \textbf{37.7} & \textbf{0.977/ 0.946} & \textbf{0.750} & \textbf{71.9} & \textbf{0.879/0.894} \\  
\bottomrule
\end{tabular}}
\end{table*}
\label{sec:ab}

According to Table~\ref{tab:ab}, when adding $AdaIN^{new}$ to the baseline, the style translation network works well for the style transfer tasks (\eg~winter$\to$summer) but is typically
unsuccessful for shape change tasks (\eg~horse$\to$zebra). The visualisation results in Figure~\ref{fig:vis_ab} (b) and (c) support this claim. Some of the semantic information belonging to the background pixel gets the completely wrong translation on the horse$\to$zebra task. However, adding the global alignment loss into the style translation network clearly provides a benefit.  
The associated qualitative results are shown in Figure~\ref{fig:vis_ab}. From the figure, better image translations are observed. {However, compared with (a) and (d) in Figure~\ref{fig:vis_ab}, there is still a little content degradation in the background. For example, the area between the two horses in the first image of Figure~\ref{fig:vis_ab} and the area around the horse's head in the second image are also striped.}

To understand how the dimension of the predicted multivariate Gaussian distributions influences the global alignment network, we perform an ablation study of the predicted multivariate Gaussian distribution dimensions. We choose $N$ sequence from $[1, 8, 16, 24, 32, 40, 48, 64]$ and test it in both Winter$\to$Summer and Horse$\to$Zebra tasks. The results are shown in Figure~\ref{fig:a}.
When the value is higher than 32, the performance of our method becomes stable. In this case, we set $N$ as 32 for all our experiments.
\subsection{Limitation}

\begin{figure}[!t]
\centering
\subfloat[Input]{
    \begin{minipage}{0.33\linewidth}
        \centering
        \includegraphics[width=0.993\textwidth,height=0.6in]{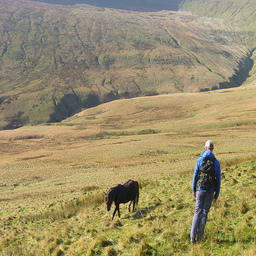}\\
        \includegraphics[width=0.993\textwidth,height=0.6in]{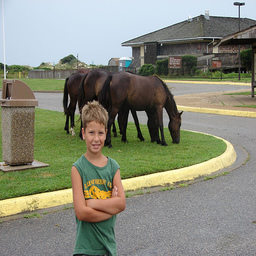}\\
        \includegraphics[width=0.993\textwidth,height=0.6in]{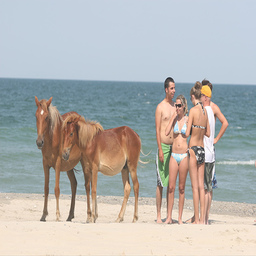}\\  
        \includegraphics[width=0.993\textwidth,height=0.6in]{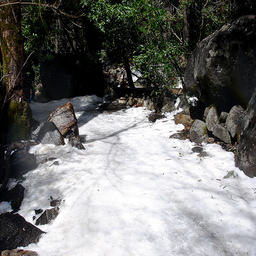}\\
        \includegraphics[width=0.993\textwidth,height=0.6in]{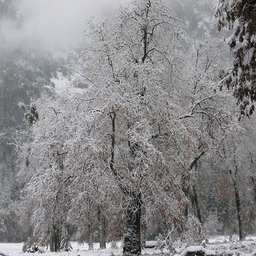}\\       
    \end{minipage}%
}%
\subfloat[Attention]{
    \begin{minipage}{0.33\linewidth}
        \centering
        \includegraphics[width=0.993\textwidth,height=0.6in]{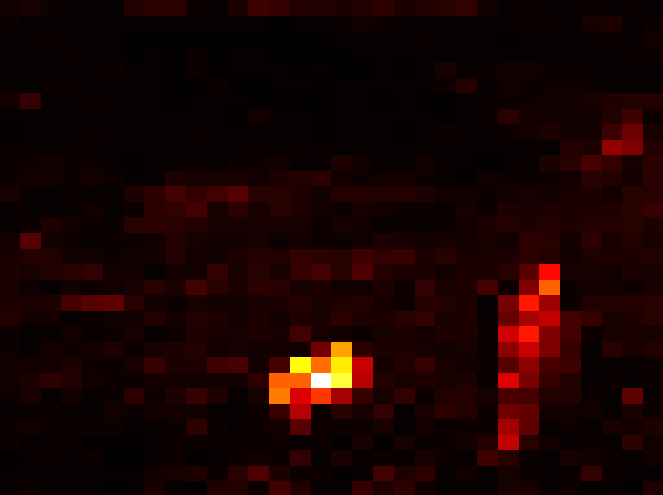}\\
        \includegraphics[width=0.993\textwidth,height=0.6in]{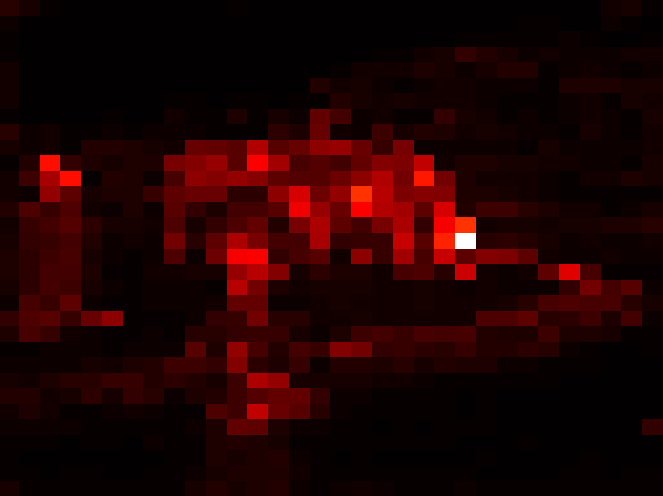}\\
        \includegraphics[width=0.993\textwidth,height=0.6in]{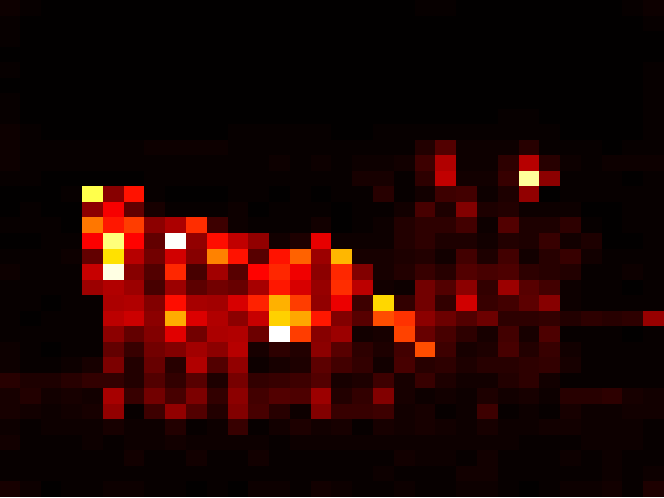}\\  
        \includegraphics[width=0.993\textwidth,height=0.6in]{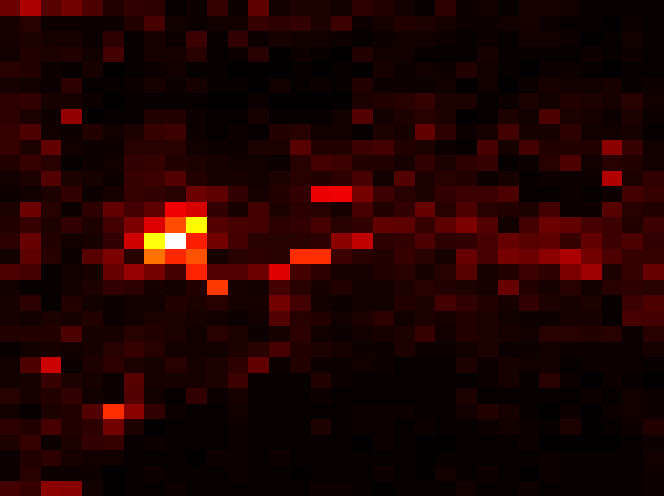}\\
        \includegraphics[width=0.993\textwidth,height=0.6in]{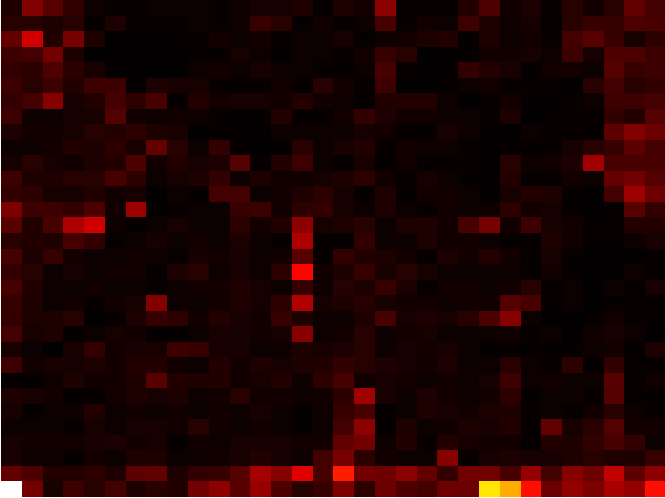}\\    
    \end{minipage}%
}%
\subfloat[Ours]{
    \begin{minipage}{0.33\linewidth}
        \centering
        \includegraphics[width=0.993\textwidth,height=0.6in]{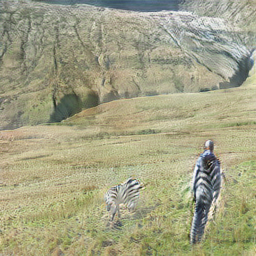}\\
        \includegraphics[width=0.993\textwidth,height=0.6in]{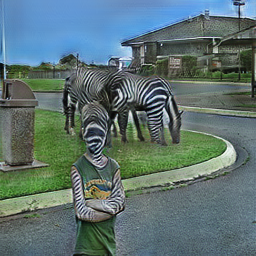}\\
        \includegraphics[width=0.993\textwidth,height=0.6in]{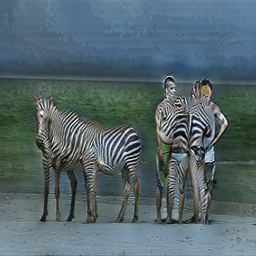}\\  
        \includegraphics[width=0.993\textwidth,height=0.6in]{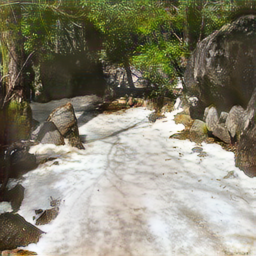}\\
        \includegraphics[width=0.993\textwidth,height=0.6in]{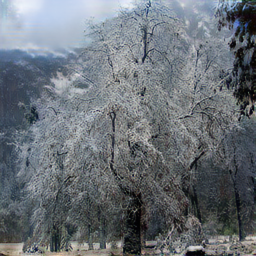}\\   
    \end{minipage}%
}%
\centering
\caption{Failure results on single-modal image translation and multi-modal image translation. }
\label{fig:vis_limit}
\vspace{-0.4cm}
\end{figure}

We show some failure results on single-modal image translation and multi-modal image translation in Figure~\ref{fig:vis_limit} to discuss our method's limitations. According to the Horse$\to$Zebra task's results, the self-attention network incorrectly provides attention when both of them coexist in the same picture. On the other hand, the failure results on the Winter$\to$Summer task prove that when the object of interest occupies a large part of the image, the self-attention network cannot identify it correctly. In the future, we plan to investigate self-distillation~\cite{caron2021emerging} to solve these problems.

\section{Conclusion}
We proposed GLA-Net, a new approach for unpaired image-to-image translation which simultaneously addresses two subtasks, \ie~style transfer and semantic content change.  Each subtask is handled by the corresponding sub-network. Specifically, we proposed a global alignment network for style transfer, which integrates an MLP-Mixer style encoder and a feature alignment strategy based on a new adaptive instance normalization scheme. 
We also introduced a local alignment network for semantic content modification, which integrates a new self-attention mechanism. We compared our approach with several state-of-the-art methods, conducting experiments on five publicly available datasets, demonstrating the superiority of our proposed GLA-Net.  

\noindent\textbf{Broader Impacts.} It is well-known that image and video generation methods based on GANs can be potentially used for malicious applications. However, focusing on the specific approach proposed here and on the application of image translation, we believe that the potential benefits, \textit{e.g.}, in creative industry tasks outweigh the threats.

\clearpage
{\small
\bibliographystyle{ieee_fullname}
\bibliography{egbib}
}

\end{document}